\documentclass[lettersize,journal]{IEEEtran}
\usepackage{amsmath,amsfonts}
\usepackage{algorithmic}
\usepackage{algorithm}
\usepackage{array}
\usepackage[caption=false,font=normalsize,labelfont=sf,textfont=sf]{subfig}
\usepackage{textcomp}
\usepackage{stfloats}
\usepackage{url}
\usepackage{verbatim}
\usepackage{graphicx}
\usepackage{cite}
\usepackage{xcolor}
\def\CHK#1 {\textcolor{magenta}{{\bf [CHK:}~#1{\bf ]}}~}
\def\ADD#1 {\textcolor{cyan}{{\bf [To add:}~#1{\bf ]}}~}
\def\qichen#1 {\textcolor{orange}{{\bf [QC:}~#1{\bf ]}}~}
\def\mj#1 {\textcolor{blue}{{\bf [MJ:}~#1{\bf ]}}~}
\def\mujian#1 {\textcolor{blue}{{\bf [MJ:}~#1{\bf ]}}~}

\def\ie{\emph{i.e.,} }

\def\etal{\emph{et al.} }
\def\Vec#1{{\boldsymbol{#1}}}

\usepackage{verbatim}
\usepackage{booktabs}

\begin{document}

\title{SHE-Net: Syntax-Hierarchy-Enhanced Text-Video Retrieval}

\author{
	Xuzheng Yu,
	Chen Jiang,
	Xingning Dong,
	Tian Gan,
	Ming Yang,
	Qingpei Guo$^{\dag}$

	\thanks{Xuzheng Yu, Chen Jiang, Xingning Dong, Ming Yang and Qingpei Guo are with the Institution of Ant Group Co., Ltd., Hangzhou 310023, China. (e-mail: $\{$yuxuzheng.yxz, qichen.jc, dongxingning.dxn, m.yang, qingpei.gqp$\}$@antgroup.com). Tian Gan are with the School of Computer Science and Technology, Shandong University, Qingdao 266237, China. (e-mail: gantian@sdu.edu.cn).}

	\thanks{Qingpei Guo$^{\dag}$ is the corresponding author.}
}




\maketitle

\begin{abstract}
The user base of short video apps has experienced unprecedented growth in recent years, resulting in a significant demand for video content analysis. In particular, text-video retrieval, which aims to find the top matching videos given text descriptions from a vast video corpus, is an essential function, the primary challenge of which is to bridge the modality gap. Nevertheless, most existing approaches treat texts merely as discrete tokens and neglect their syntax structures. Moreover, the abundant spatial and temporal clues in videos are often underutilized due to the lack of interaction with text. To address these issues, we argue that using texts as guidance to focus on relevant temporal frames and spatial regions within videos is beneficial. In this paper, we propose a novel \textbf{S}yntax-\textbf{H}ierarchy-\textbf{E}nhanced text-video retrieval method \textit{(SHE-Net)} that exploits the inherent semantic and syntax hierarchy of texts to bridge the modality gap from two perspectives. First, to facilitate a more fine-grained integration of visual content, we employ the text syntax hierarchy, which reveals the grammatical structure of text descriptions, to guide the visual representations. Second, to further enhance the multi-modal interaction and alignment, we also utilize the syntax hierarchy to guide the similarity calculation. We evaluated our method on four public text-video retrieval datasets of MSR-VTT, MSVD, DiDeMo, and ActivityNet. The experimental results and ablation studies confirm the advantages of our proposed method.
\end{abstract}

\begin{IEEEkeywords}
Information Search and Retrieval, Text-Video Retrieval, Syntax-Hierarchy-Enhanced, Multimodal Fusion and Embedding.
\end{IEEEkeywords}

\section{INTRODUCTION}
    \IEEEPARstart{I}{n} recent years, short video apps have undergone unprecedented growth, resulting in a high demand for video content analysis and reasoning. The primary demand for video media software is to expeditiously search for relevant videos that meet user intentions from a vast video corpus. To this end, the text-video retrieval task~\cite{zhu2023deep, TS2_Net, CLIP4clip} is a natural and intuitive solution, which has raised increasing research interests in recent years. The goal of this task is to find the top matching videos according to given text descriptions, where the critical challenge is to bridge the modality gap between texts and videos.

\begin{figure*}[t]
        \centering
        \includegraphics[width=0.8\textwidth,height=0.3\textwidth]{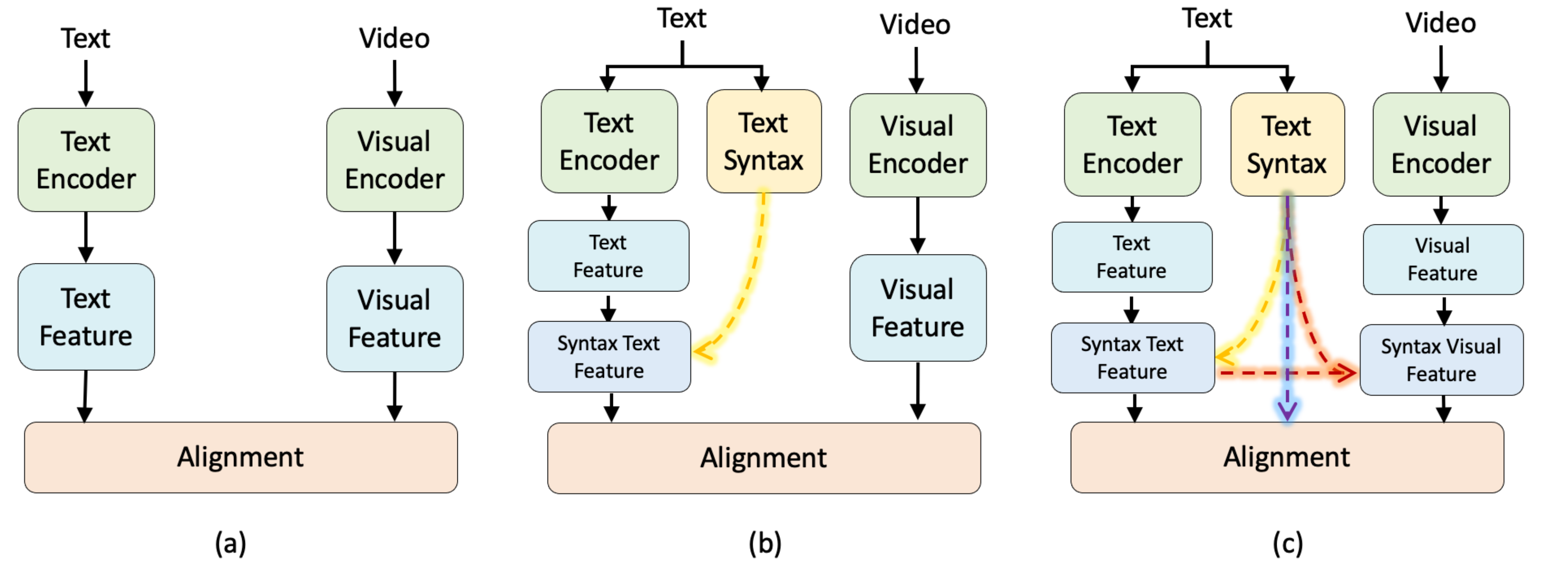} 
        \vspace{-0.4cm}
        \caption{The illustration of different ways of using syntax hierarchical structures. Among them, (a) illustrates the traditional twin-tower structure of text-video retrieval models, and (b) represents the prior syntax-guided approaches that predominantly use syntax structures to bolster text features, potentially neglecting video modeling. In our proposed method (c), we develop the text syntax hierarchy and leverage it to assimilate the grammatical structure of the given text, we also construct the video syntax hierarchy under the guidance of the text syntax one, and design a similarity calculation method guided by the text syntax structure.}
        \vspace{-0.4cm}
        \label{method_new_intro}
    \end{figure*}

    Most recent text-video retrieval methods~\cite{CLIP4clip,CLIP2TV,DRL,TS2_Net,X_CLIP,STAN,DMAE} follow a typical pipeline, which can be divided into three modules: text encoding, video encoding, and text-video alignment, as shown in Fig.\ref{method_new_intro}(a). Specifically, given a mini-batch of video-text pairs, most methods initially extract text and video features with specialized uni-modal encoders, and subsequently feed them into the text-video alignment module for learning cross-modal similarities. 
    These methods typically treat texts and videos as discrete tokens simply. Yet texts and videos constitute two inherently different modalities, exhibiting a natural asymmetry in grammatical structures.
    Concretely, text inputs are grammatically well structured, while video data, being spatial-temporal in nature, often contains a significant amount of redundancy and lacks the structured grammar found in texts. 
    Hence, it remains a very challenging issue to effectively model cross-modal similarities at various granularities. 
    Some works~\cite{HANet, FGAR, FGVTR} have attempted to introduce the text syntax into the text encoding module, as shown in Fig.\ref{method_new_intro}(b), yet potentially neglecting video modeling and cross-modal interaction.
    To alleviate this problem, we propose a novel \textbf{S}yntax-\textbf{H}ierarchy-\textbf{E}nhanced method \textit{(SHE-Net)} that exploits the inherent semantic and syntax hierarchy of texts to bridge the modality gap. Different from the previous works that only introduce the text syntax into the text encoding module, we further adopt it as a guidance for the video encoding and text-video alignment module to facilitate better visual representations and enhance the multi-modal interaction, as shown in Fig.\ref{method_new_intro}(c).

\begin{figure*}[t]
    \centering
    \includegraphics[width=0.9\textwidth,height=0.55\textwidth]{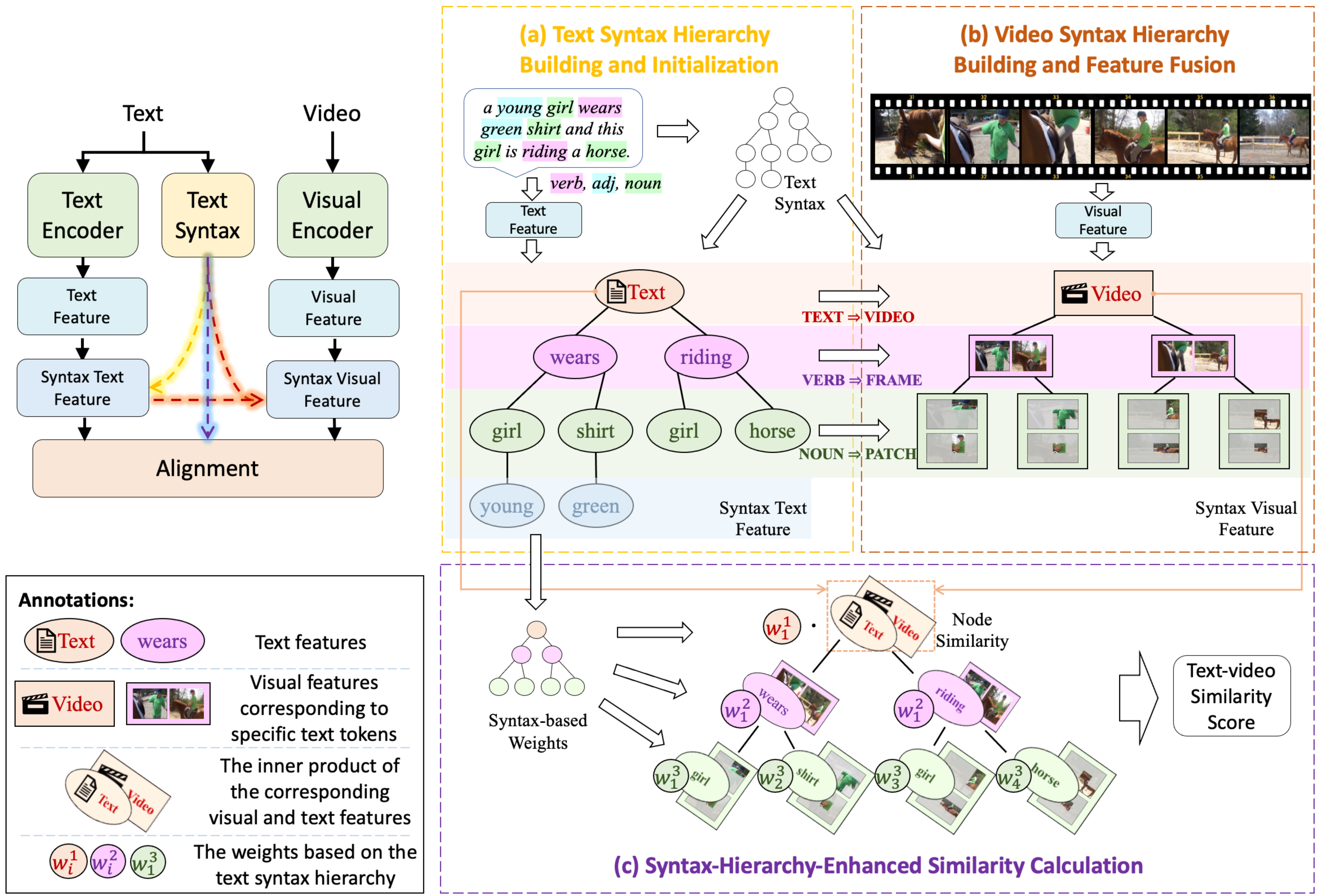}
    \vspace{-0.3cm}
    \caption{The overview of the proposed SHE-Net method and main contributions. (a) SHE-Net first constructs and utilizes the text syntax hierarchy to learn the grammatical structure of the given text. (b) This method then develops the video syntax hierarchy, guided by the corresponding text syntax, effectively aggregating numerous video signals into information-dense. (c) Finally, by leveraging these two syntax hierarchies jointly, this method enhances the cross-modal alignment across various levels of granularity, thus achieving superior retrieval performance.}
    \vspace{-0.4cm}
    \label{fig:teaser}
\end{figure*}

    \IEEEpubidadjcol
    First, regarding text encoding, we advocate for the explicit exploitation of the grammatical structure of the given text descriptions. Intuitively, humans tend to focus on certain keywords and the overall syntactic structure to understand a sentence. For example, certain words (\emph{e.g.}, verbs, nouns, and adjectives) provide the essential information and whose positions determine the meaning of the entire sentence. However, most existing text-video retrieval methods do not explicitly exploit this informative syntax structure.
    They mainly either 1) encode the entire text description holistically~\cite{CLIP4clip, CLIP2Video, Frozen_in_Time}, unable to utilize the relation among the word embeddings further, or
    2) merely assign more attention to keywords within a sentence~\cite{HANet, FGAR, FGVTR}, without leveraging the rich and valuable syntax structure.
    Towards this end, as illustrated in Fig.~\ref{fig:teaser}(a), we first employ an off-the-shelf abstract-syntax-tree toolkit~\cite{SPACY} to generate a syntax hierarchy structure for each input text, which is then incorporated into the text encoding module to formally model the rich grammatical structure. In short, the constructed text syntax hierarchy aims to emphasize those keywords within a given text, thereby extracting the syntax information for enhanced text encoding.

    Second, for visual encoding, we propose aggregating abundant video signals into information-dense video features. 
    Compared with texts, which comprise human-generated signals, videos constitute natural signals with substantial spatial and temporal redundancy, which could be significantly compressed. Most text-video retrieval methods do not take advantage of this intrinsic characteristic. They typically 1) retain or cluster all video tokens during the feature extraction step~\cite{DRL, TS2_Net, X_CLIP}, thereby introducing irrelevant or redundant information; or 2) resort to obtaining a single token through pooling~\cite{CLIP4clip}, resulting in a loss of the spatial-temporal clues.
    To address this issue, we seek to condense massive low-level video signals into a compact set of information-rich features under the guidance of textual cues. Specifically, we leverage the constructed text syntax hierarchy to establish a corresponding video syntax one, which is incorporated into the video encoding module to derive information-rich video features. As illustrated in Fig.~\ref{fig:teaser}(b), we employ each \textbf{verb} within the text syntax hierarchy to retrieve a fixed number of \textbf{frames} in the video syntax hierarchy, following the intuition that a verb typically signifies an action occurring within a certain period of time. Additionally, we map each \textbf{noun} to corresponding image regions or \textbf{patches}, reflecting the notion that a noun typically represents a static entity that can be localized within a frame. In summary, by utilizing the text syntax hierarchy to guide the construction of the video syntax structure, we can effectively filter out irrelevant video signals and extract information-rich video features.

    Furthermore, to align text and video embeddings, we propose designing a syntax hierarchy enhanced multi-granularity cross-modal fusion strategy to enhance the text-video retrieval performance. Some text-video retrieval methods use pooled text and video features to calculate global cross-modal similarity scores~\cite{CLIP4clip, FGAR}. 
    Alternatively, other methods extract distinct tokens from the entire feature sequence to perform unguided local cross-modal similarity calculations at various granularities~\cite{DRL, FGVTR}. 
    It offers a significant advantage to conduct text-video alignment at various granularities, guided by semantic and grammatical structures. Therefore, as illustrated in Fig.~\ref{fig:teaser}(c), we leverage the constructed text and video syntax hierarchies to conduct cross-modal alignment at multiple levels of granularity. Specifically, in addition to the global ``text-video'' alignment, we establish the ``verb-frame" alignment to model the temporal correspondence between video-text pairs, as well as the ``noun-patch/region'' alignment to capture the spatial congruence.
    In this way, by employing these corresponding syntax hierarchies, we extract rich spatio-temporal cues across different granularities, thereby effectively enhancing the text-video retrieval performance.

    In summary, as illustrated in Fig.~\ref{fig:teaser}, we first construct and utilize the text syntax hierarchy to learn the grammatical structure of the given text. We then develop the video syntax hierarchy, guided by the corresponding text syntax, effectively aggregating numerous video signals into information-dense video features. Finally, by leveraging these two syntax hierarchies jointly, we enhance the cross-modal alignment across various levels of granularity, thus achieving superior retrieval performance. Our approach optimizes all three components in the general pipeline, mitigating some of the aforementioned limitations in the current text-video retrieval task.

    Our main contributions are summarized as three-folds:
    \begin{itemize}
        \item We propose a novel \textbf{S}yntax-\textbf{H}ierarchy-\textbf{E}nhanced text-video retrieval method, dubbed \textit{SHE-Net}, which capitalizes on the inherent semantic and syntax structure of texts to guide the integration of visual content and calculation of the text-video similarity.
        
        \item We first construct and extend the text syntax hierarchy to capture generalized grammatical structures in texts, and then establish a video syntax hierarchy guided by the text syntax hierarchy, which enables the model to select the most relevant video tokens from massive candidates. Finally, we augment multi-modal interaction and alignment by leveraging both hierarchies together.
        
        \item We conduct a quantitative evaluation of our \textit{SHE-Net} on four public text-video retrieval benchmarks, including MSR-VTT, MSVD, DiDeMo and ActivityNet. Experimental results demonstrate the effectiveness of our proposed method.
    \end{itemize}

\section{RELATED WORK}



\subsection{Vision-Language Pre-Training}
    Vision-language models~\cite{WenLan, ALBEF, CLIP, Florence}
    pretrained on large-scale image/video-text datasets ~\cite{VideoBERT, HERO, Frozen_in_Time, HowTo100M, SNPS3, VSFT, STVLB}
    have shown excellent performance in recent years.
    Existing vision-language pre-training methods could be roughly divided into three categories, \ie{shared-encoder-based methods, multi-encoders-fusion methods, and two-stream-style methods}. Shared-encoder-based methods exploit~\cite{VL_BERT, Unicoder_VL} a shared encoder to embed text and visual sequences. Multi-encoders-fusion ones~\cite{HERO, Univl, MMT} generally leverage three separate encoders to model visual, textual, and cross-modal features. While two-stream-style ones~\cite{Frozen_in_Time, CLIP} could be treated as a twin-tower architecture, containing two parallel encoders to encode textual and visual features with a contrastive loss to learn cross-modal alignment. Among all the aforementioned methods, CLIP~\cite{CLIP}, a two-stream-style model pre-trained on more than 400M image-text pairs, is one of the most widely-adopted models due to its superior performance. Therefore, In our work, we adopt CLIP as our basic architecture.

\vspace{-0.1cm}
\subsection{CLIP-based Text-Video Retrieval}
    The remarkable success of CLIP~\cite{CLIP} has proven effective in numerous downstream tasks, including image-text understanding~\cite{COOP, DWAN, ILGIT}, 
    video action recognition~\cite{ActionCLIP, AGPN}, video understanding~\cite{LTVQA, BAQIVQ, ACRTN, HMMAN, CERN, EBRAM, MLGAN}, etc. CLIP has notably contributed to the advancement of video retrieval tasks, as evidenced by recent works like CLIP4Clip~\cite{CLIP4clip}, TS2-Net~\cite{TS2_Net}, STAN~\cite{STAN}, MULTI~\cite{MULTI} and others ~ \cite{RSIVR, TMGT, UMCKD, CASLVTR}.
    Videos, unlike images, contain abundant spatio-temporal information, presenting unique challenges for adaptation. While numerous methods, including CLIP4clip~\cite{CLIP4clip}, DRL~\cite{DRL}, X-CLIP~\cite{X_CLIP} and others, attempt to enhance the CLIP architecture with temporal modelling, they typically yield only incremental improvements. Consequently, their ability to generate robust video representations remains constrained. Towards this end, CLIP-ViP~\cite{CLIP_VIP} incorporates temporal signals into the visual encoder to better integrate the rich spatio-temporal information. Different from simply leveraging the CLS token to serve as frame features, CenterCLIP~\cite{CenterCLIP}, TS2-Net~\cite{TS2_Net}, etc. introduce the remaining token features to achieve more informative and robust video representations. Unlike the aforementioned methods, in our work, we abstract the massive redundant video tokens into a visual syntax hierarchy with the guidance of text descriptions, and then exploit it to enhance the quality of generated multi-modal features.
    

\vspace{-0.1cm}
\subsection{Text-Guided Text-Video Retrieval}
    Text-Video Retrieval is essentially a cross-modal alignment task requiring sufficient multi-modal interaction. Therefore, unlike the traditional methods~\cite{CLIP4clip, CenterCLIP, CLIP_VIP}
    that separately generate video and text features without any interaction except for the final similarity calculation, many text-guided fusion strategies~\cite{DRL, FGAR, HANet, FGVTR}
    have been proposed to perform cross-modal interaction at an earlier stage.
    Under this strategy, even for the same video, different query texts can focus on different parts of the video, and can guide the model to obtain different query-related features.
    For example, Gorti~\etal\cite{X_Pool} highlighted the drawbacks of text-agnostic video pooling and present an alternative framework for text conditioned pooling. 
    Besides, Wu~\etal\cite{HANet} proposed the hierarchical alignment network to make full use of complementary information of different semantic levels of representations. 
    However, these methods generally only utilize syntax structure to enhance text features, which may be insufficient in video modeling.
    Moreover, unlike languages that are human-generated signals with rich semantics, videos are natural signals with heavy spatial and temporal redundancy.
    Unfortunately, current text-guided strategies do not exploit this intrinsic characteristic explicitly.
    Towards this end, we first establish a textual syntax hierarchy to highly abstract the input text description, and employ it to build a corresponding video syntax hierarchy. With the guidance of these twin parallel syntax hierarchies, we could obtain more informative multi-modal video representations and achieve better text-video retrieval performance.

\section{OUR PROPOSED METHOD}
\subsection{Overview}
    \subsubsection{Problem setting.} 
        Formally, let $\mathcal{V}$ and $\mathcal{T}$
        represent the set of videos and text descriptions, respectively.
        Each video $v \in \mathcal{V}$ corresponds to one or multiple textual descriptions, and comprises a sequence of RGB frames $\{\Vec{I}_{1}, \Vec{I}_{2}, ..., \Vec{I}_{N_{\rm{v}}}\}$, where $N_{\rm{v}}$ denotes the number of the sampled frames.
        Our objective is to develop a text-video retrieval model capable of finding the most matching videos corresponding to the textual descriptions.
    \vspace{-0.1cm}
    \subsubsection{Model overview.}
        Videos are natural signals with significant spatial and temporal redundancy, whereas text descriptions are concise and precise human-generated signals.
        Inspired by this observation, we argue that text can guide the selection of video segments containing relevant information, mitigating the negative impact of redundant or even noisy visual signals.
        In this paper, we first build the text syntax tree to generalize the grammar structure of the given text, and convert this tree into a text syntax hierarchy architecture. Moreover, we also establish a video syntax hierarchy guided by the text syntax hierarchy, enabling the model to select the most informative video tokens from massive redundant candidates. Ultimately, we exploit these two syntax hierarchies to jointly enhance the multi-modal interaction and alignment.
\vspace{-0.1cm}

\subsection{Text Syntax Hierarchy Building and Initialization}
    Recent studies \cite{HANet, FGAR, FGVTR} have confirmed that the rich structural information in text descriptions significantly aids text-video retrieval tasks. These methods typically partition textual descriptions into three levels: whole, action, and entity. The whole level describes the overall content of videos, including various independent actions across temporal frames, and these actions involve diverse entities. This hierarchical, global-to-local structure aids in a more nuanced understanding of the text content. In this work, we adopt this hierarchical approach. Additionally, we extend our model to incorporate a fourth layer (\ie{the adjective layer}) of entity description based on the three-layer hierarchical structure, which is employed to describe entities with various attributes (such as color) in a more fine-grained manner.

    Specifically, given a text description $\Vec{t}=\{q_1, q_2, ..., q_{N_{\rm{t}}}\}$ containing $N_{t}$ words, we apply an off-the-shelf syntax analysis toolkit to obtain the parts-of-speech tag of each word, and grammatical dependencies among words. We then build a syntax tree according to the grammatical dependencies, and parse this syntax tree into a four-level syntax hierarchical (\ie{the sentence, verbs, nouns, and adjectives}) structure, which follows a global-to-local architecture.
    In the obtained four-layer hierarchical structure, the first layer $\Vec{H}^1$ has only one special overall node $\Vec{h}^{t1}_1$, representing the whole text. The second layer $\Vec{H}^2=\{\Vec{h}^{2}_1, \Vec{h}^{2}_2, ..., \Vec{h}^{2}_{|\Vec{H}^2|}\}$ contains all verbs in sentences, 
    and all verb nodes are connected to the overall node. Subsequently, the third layer $\Vec{H}^3=\{\Vec{h}^{3}_1, \Vec{h}^{3}_2, ..., \Vec{h}^{3}_{|\Vec{H}^3|}\}$ contains all entity nouns, which describe the entities involved in actions, and each node in $\Vec{H}^{3}$ is connected to its associated verb nodes in $\Vec{H}^{2}$. Specially, in instances where verbs are absent or when the analysis tools cannot establish these relationships, we introduce a special token, $[$EXIST$]$. This token symbolizes the notion of existence and links it to all words in the sentence. This mechanism helps maintain the model’s functional efficacy by preserving semantic coherence, even in incomplete syntactic structures. Finally, the bottom fourth layer $\Vec{H}^4=\{\Vec{h}^{4}_1, \Vec{h}^{4}_2, ..., \Vec{h}^{4}_{|\Vec{H}^4|}\}$ is the adjectives, connecting to their related entity noun nodes. 

    After obtaining the above hierarchical structure, we first apply the text encoder of CLIP \cite{CLIP} $\emph{TextEnc}$ to encode the text description with the special $[$CLS$]$ token $\Vec{q}_{\rm{cls}}$, and obtain the feature sequence $\Vec{F}^{\rm{t}}$ as follows:
    \begin{equation}
    \setlength{\abovedisplayskip}{2pt}
    \setlength{\belowdisplayskip}{2pt}
    \begin{split}
    &\ \ \ \ \ \ \Vec{F}^{\rm{t}} = \left[\Vec{f}^{\rm{t}}_{\rm{cls}}, \Vec{f}^{\rm{t}}_{1}, \Vec{f}^{\rm{t}}_{2}, ... , \Vec{f}^{\rm{t}}_{N_{\rm{t}}}\right]\\
    &=\emph{TextEnc}([ \Vec{q}_{\rm{cls}}, \Vec{q}_{1}, \Vec{q}_{2}, ... , \Vec{q}_{N_{\rm{t}}}])
    \end{split},
    \end{equation}
    where $N_{\rm{t}}$ indicates the number of words in the query text, and $\Vec{q}_{\rm{cls}}$ represents the special $[$CLS$]$ token.

    Subsequently, for nodes in the hierarchical structure, we select their corresponding token features from the feature sequence $\Vec{F}^{\rm{t}}$ as their initial features. We denote the initial features of nodes in each hierarchical layer $\Vec{H}^{\theta}$ as $\Vec{f}^{\rm{t}\theta}_{i}$, where $\theta \in \left[1, 2, 3, 4\right]$ represents the level in the syntax hierarchy. Specially, we utilize the [CLS] feature of text descriptions as the initial overall node feature $\Vec{f}^{\rm{t1}}_{1}$. The entire feature initialization step for the text syntax hierarchy could be formulated as:
    \begin{equation}
    \setlength{\abovedisplayskip}{2pt}
    \setlength{\belowdisplayskip}{2pt}
    \Vec{f}^{\rm{t}\theta}_{i}=\begin{cases}
    \Vec{f}^{\rm{t}}_{\rm{cls}} &\theta \in \left[1\right]  \\
    \Vec{f}^{\rm{t}}_{\mu\left(\Vec{h}^{\theta}_{i}\right)} &\theta \in \left[2, 3, 4\right]
    \end{cases},
    \end{equation}
    where $\mu\left(\cdot\right)$ indicates the position of a word in the text description.

    After that, we utilize the adjective nodes in $\Vec{H}^{4}$ to enhance the entity noun nodes in $\Vec{H}^{3}$ with the attention mechanism, and obtain the enhanced entity node features $\Vec{f}^{\rm{t3'}}_{i}$ as follows:
    \begin{equation}
    \setlength{\abovedisplayskip}{2pt}
    \setlength{\belowdisplayskip}{1pt}
    \Vec{e}^{\rm{t3'}}_{i}=\operatorname{Norm}\left(\Vec{f}^{\rm{t3}}_{i} + \operatorname{MLP}_{4}\left(\ \Vec{f}^{\rm{t3}}_{i} \right)\right),
    \end{equation}
    \begin{equation}
    \setlength{\abovedisplayskip}{1pt}
    \setlength{\belowdisplayskip}{1pt}
    \alpha^{\rm{desc}}_{i,j}=\tfrac{\operatorname{exp}\left(\Vec{e}^{\rm{t3'}}_{i}\odot\Vec{f}^{\rm{t4}}_{j}\right)}{\sum_{j\in \varphi^{\rm{3}}_{i}}\operatorname{exp}\left(\Vec{e}^{\rm{t3'}}_{i}\odot\Vec{f}^{\rm{t4}}_{j}\right)},
    \end{equation}
    \begin{equation}
    \setlength{\abovedisplayskip}{1pt}
    \setlength{\belowdisplayskip}{0pt}
    {\gamma}^{\rm{desc}}_{i}=\sum_{j\in \varphi^{\rm{3}}_{i}} \alpha^{\rm{desc}}_{i,j}\Vec{f}^{\rm{t4}}_{j},
    \end{equation}
    \begin{equation}
    \setlength{\abovedisplayskip}{1pt}
    \setlength{\belowdisplayskip}{2pt}
    \Vec{f}^{\rm{t3'}}_{i}=\Vec{e}^{\rm{t3'}}_{i}+\operatorname{Fusion}\left(\Vec{e}^{\rm{t3'}}_{i}\oplus{\gamma}^{\rm{desc}}_{i}\right),
    \end{equation}
    where $\operatorname{Norm}\left(\cdot\right)$ indicates the layer normalization operation, $\operatorname{MLP}_{4}\left(\cdot\right)$ indicates the multilayer perceptron (MLP) mapped from d-dimensions to d-dimensions, $\varphi^{\rm{\theta}}_{i}$ represents the set of nodes in the $(\theta+1)$-th layer adjacent to the node $\Vec{h}^{\theta}_{i}$ in the $\theta$-th layer, $\odot$ indicates the operation of inner product, $\operatorname{Fusion}\left(\cdot\right)$ indicates the multilayer perceptron compressed from 2d-dimensions to d-dimensions and $\oplus$ indicates the operation of concatenating.

\subsection{Video Syntax Hierarchy Building and Feature Fusion}

    Videos are natural signals with significant spatial and temporal redundancy. In existing literature~\cite{FGVTR, HANet}, the methods for obtaining video features have typically utilized separate multilayer perceptron (MLP) encoders to extract features at the video, action, and entity levels from frame-level features, which have limited capability in aggregating informative video features. In this paper, we propose leveraging the informative and hierarchical structure of text to guide the fusion of video features at varying granularities. Considering that videos exhibit a hierarchical structure analogous to text, we aim to enhance video representations and filter out redundant signals across granularity levels. To this end, we decompose videos into three hierarchical components, namely video content, frames (i.e., actions), and patches (i.e., entities), establishing a three-tier video syntax hierarchy in the sequence of ``video content-actions-entities". Through this approach, we employ distinct analytical methods tailored to each granularity level to enhance the extracted features.

    Specifically, as the basis for the feature fusion stage, we first apply the visual encoder of CLIP \cite{CLIP} $\emph{VisEnc}$ to encode the basic visual features. Given a sampled frame $\Vec{I}_{i}$, we split it into $N_p$ patches $\{\Vec{p}_{i,1}, \Vec{p}_{i,2}, ... , \Vec{p}_{i,{N_p}}\}$, and then feed them into transformer blocks with the $[$CLS$]$ token $\Vec{p}_{\rm{cls}}$ as follows:
    \begin{equation}
    \setlength{\abovedisplayskip}{2pt}
    \setlength{\belowdisplayskip}{2pt}
    \begin{split}
    &\ \ \ \ \ \ \ \Vec{F}^{\rm{v}}_{i} = \left[\Vec{f}^{\rm{v}}_{i,\rm{cls}}, \Vec{f}^{\rm{v}}_{i,1}, \Vec{f}^{\rm{v}}_{i,2} ..., \Vec{f}^{\rm{v}}_{i,N_{\rm{p}}}\right]\\
    &=\emph{VisEnc}([ \Vec{p}_{\rm{cls}}, \Vec{p}_{i,1}, \Vec{p}_{i,2}, ... , \Vec{p}_{i,N_{\rm{p}}}])
    \end{split},
    \end{equation}
    where $\Vec{f}^{\rm{v}}_{i,\rm{cls}}$ is the obtained frame-level feature of the frame $I_i$, and $\Vec{f}^{\rm{v}}_{i,j}$ represents the obtained local features corresponding to the patch $\Vec{p}_{i,j}$. 


    Subsequently, for the overall node $\Vec{h}^{1}_{1}$ in the first layer $\Vec{H}^{1}$, we directly utilize the text feature of this node to guide the fusion of all frame-level visual features, obtaining text-dependent global video representations $\Vec{e}^{\rm{v1}}_{1}$ as follows:
    \begin{equation}
    \setlength{\abovedisplayskip}{2pt}
    \setlength{\belowdisplayskip}{0pt}
    \Vec{e}^{\rm{t1}}_{1}=\operatorname{Norm}\left(\Vec{f}^{\rm{t1}}_{1} + \operatorname{MLP}_{1}\left(\ \Vec{f}^{\rm{t1}}_{1} \right)\right),
    \end{equation}
    \begin{equation}
    \setlength{\abovedisplayskip}{0pt}
    \setlength{\belowdisplayskip}{0pt}
    \alpha^{\rm{cls}}_{1,j}=\tfrac{\operatorname{exp}\left(\Vec{e}^{\rm{t1}}_{1}\odot\Vec{f}^{\rm{v}}_{j,\rm{cls}}\right)}
    {\sum_{j\in [1,N_{\rm{v}}]}\operatorname{exp}\left(\Vec{e}^{\rm{t1}}_{1}\odot\Vec{f}^{\rm{v}}_{j,\rm{cls}}\right)},
    \end{equation}
    \begin{equation}
    \setlength{\abovedisplayskip}{0pt}
    \setlength{\belowdisplayskip}{0pt}
    \Vec{e}^{\rm{v1}}_{1}=\sum_{j\in [1,N_{\rm{v}}] } \alpha^{\rm{cls}}_{1,j}\Vec{f}^{\rm{v}}_{j,\rm{cls}}.
    \end{equation}
    where $\odot$ indicates the operation of inner product, and $N_{\rm{v}}$ represents the number of the sampled frames.

    Besides, for each action node in the second layer, we first apply Transformer to fuse the temporal information to obtain temporal-encoded features $\Vec{g}^{\rm{v}}$ of each frame, and then we utilize the text features of action nodes to guide the fusion of the temporal-encoded features of the selected frames, obtaining features $\Vec{e}^{\rm{v2}}_{i}$ related to specific actions $\Vec{h}^{2}_i$ in videos as follows:
    \begin{equation}
    \setlength{\abovedisplayskip}{2pt}
    \setlength{\belowdisplayskip}{0pt}
    \Vec{e}^{\rm{t2}}_{i}=\operatorname{Norm}\left(\Vec{f}^{\rm{t2}}_{i} + \operatorname{MLP}_{2}\left(\ \Vec{f}^{\rm{t2}}_{i} \right)\right),
    \end{equation}
    \begin{equation}
    \setlength{\abovedisplayskip}{0pt}
    \setlength{\belowdisplayskip}{0pt}
    \left[\Vec{g}^{\rm{v}}_{1}, ..., \Vec{g}^{\rm{v}}_{N_{\rm{v}}}\right] =
    \operatorname{Transformer}([\Vec{f}^{\rm{v}}_{1,\rm{cls}}, ... , \Vec{f}^{\rm{v}}_{N_{\rm{v}},\rm{cls}}]),
    \end{equation}
    \begin{equation}
    \setlength{\abovedisplayskip}{0pt}
    \setlength{\belowdisplayskip}{0pt}
    \Vec{\psi}^{2}_{i}=argTopK^{{\lambda}_{\rm{frame}}}_{j}\left(\{\Vec{e}^{\rm{t2}}_{i}\odot\Vec{g}^{\rm{v}}_{j} \mid  j\in \left[ 1,N_{\rm{v}}\right]\}\right),
    \end{equation}
    \begin{equation}
    \setlength{\abovedisplayskip}{0pt}
    \setlength{\belowdisplayskip}{2pt}
    \Vec{e}^{\rm{v2}}_{i}=\tfrac{1}{{\lambda}_{\rm{frame}}}\sum_{j\in \Vec{\psi}^{2}_{i}}\Vec{g}^{\rm{v}}_{j},
    \end{equation}
    where $argTopK^{k}_{j}\left({\rm{val}_j}\right)$ indicates the set of parameters $j$ corresponding to the Top-$k$ values in the set $\{\rm{val}_{j}\}$, $\Vec{\psi}^{2}_{i}$ represents the collection of frames selected by action nodes, and ${\lambda}_{\rm{frame}}$ is a hyperparameter that controls how many frames to select.

    Moreover, for the enhanced entity noun nodes in $\Vec{H}^{3}$, we utilize them to guide the fusion of the selected patches, which contain more fine-grained local information than CLS features. Since different entity noun nodes in the hierarchy are only adjacent to specific action nodes, we only utilize the patch features in the specific frames selected by those adjacent action nodes as follows:
    \begin{equation} 
    \setlength{\abovedisplayskip}{2pt}
    \setlength{\belowdisplayskip}{0pt}
    \Vec{e}^{\rm{t3}}_{i}=\operatorname{Norm}\left(\Vec{f}^{\rm{t3'}}_{i} + \operatorname{MLP}_{3}\left(\Vec{f}^{\rm{t3'}}_{i} \right)\right),
    \end{equation}
    \begin{equation} 
    \setlength{\abovedisplayskip}{0pt}
    \setlength{\belowdisplayskip}{0pt}
    \Vec{\psi}^{3}_{i,j}=argTopK^{{\lambda}_{\rm{patch}}}_{x}\left(\{\Vec{e}^{\rm{t3}}_{i}\odot\Vec{f}^{\rm{v}}_{j,x} \mid  x\in \left[ 1,N_{p}\right]\}\right),
    \end{equation}
    \begin{equation} 
    \setlength{\abovedisplayskip}{0pt}
    \setlength{\belowdisplayskip}{0pt}
    \Vec{e}^{\rm{v3}}_{i,j}=\tfrac{1}{{\lambda}_{\rm{patch}}}\sum_{x\in \Vec{\psi}^{3}_{i,j}}\Vec{f}^{\rm{v}}_{j,x},
    \end{equation}
    \begin{equation} 
    \setlength{\abovedisplayskip}{0pt}
    \setlength{\belowdisplayskip}{0pt}
    \Vec{e}^{\rm{v3}}_{i}=\tfrac{1}{{\lambda}_{\rm{patch}}}\sum_{j\in \Vec{\psi}^{2}_{\delta^{3}_{i}}} \Vec{e}^{\rm{v3}}_{i,j},
    \end{equation} where $argTopK^{k}_{j}\left({\rm{val}_j}\right)$ indicates the set of parameters $j$ corresponding to the Top-$k$ values in the set $\{\rm{val}_{j}\}$, $\Vec{\psi}^{3}_{i,j}$ represents the collection of patches selected by the enhanced entity noun nodes $\Vec{h}^{3}_i$ in the $j$-th frame, $\delta^{\theta}_{i}$ represents the node in the $(\theta-1)$-th layer adjacent to node $\Vec{h}^{\theta}_{i}$ in the $\theta$-th layer, and ${\lambda}_{\rm{patch}}$ is a hyperparameter that controls how many patches to select in each picked frame.

    So far, for each node in the first three layers of the text hierarchy, we apply layer-related methods to obtain the text-guided hierarchical visual features, which also contain three layers and correspond to the text hierarchy one by one.

\vspace{-0.1cm}
\subsection{Syntax-Hierarchy-Enhanced Similarity Calculation}
    After obtaining the visual hierarchy corresponding to the text hierarchy, we design a syntax-hierarchy-enhanced similarity computation to obtain the final text-video similarity. Unlike previous work, which usually calculates the similarity between global-pooling multi-modal features or roughly calculates the similarity between the corresponding layers, we first align the established text and video syntax hierarchical nodes and calculate their similarity scores node-by-node as follows:
    
    \begin{equation}
    {score}^{\theta}_{i}=\Vec{e}^{\rm{t}\theta}_{i}\odot\Vec{e}^{\rm{v}\theta}_{i},
    \end{equation}
    where $\theta \in \left[1, 2, 3\right]$ represents the level in the syntax hierarchy.

    \begin{table*}[]
    \centering
    \caption{Cross-modal retrieval comparison with state-of-the-art methods on MSR-VTT. The symbol $\ast$ indicates the usage of DSL\cite{CAMoE} post-processing operation in methods, whereas the symbol $\dag$ denotes that the results are obtained through our re-training process to align the setting of features and environments.
    }
    \vspace{-0.2cm}
    \label{tab:res_msrvtt}
    \renewcommand{\arraystretch}{0.8}
    \setlength\tabcolsep{2pt}
    \scalebox{1.0}{
    \begin{tabular}{l| c c c c c c| c c c c c c}
        \toprule
       \multicolumn{1}{c}{} & \multicolumn{6}{|c}{Text-to-Video Retrieval} & \multicolumn{6}{|c}{Video-to-Text Retrieval} \\ \toprule
        {Method} & {R$@$1$\uparrow$} & {R$@$5$\uparrow$} & {R$@$10$\uparrow$} & {MdR$\downarrow$} & {MeanR$\downarrow$} & {Rsum$\uparrow$} & {R$@$1$\uparrow$} & {R$@$5$\uparrow$} & {R$@$10$\uparrow$} & {MdR$\downarrow$} & {MeanR$\downarrow$} & {Rsum$\uparrow$} \\ \toprule
        {CLIP4Clip\cite{CLIP4clip}} &44.5 &71.4 &81.6 &2.0  &15.3 &197.5 &- &- &- &-&- &- \\ 
        {CenterCLIP\cite{CenterCLIP}} & 44.2  &71.6 &82.1 &2.0  &15.1 &197.9 &42.8  &71.7 &82.2 &2.0  &10.9 &196.7 \\
        {HGR$^\dag$\cite{FGVTR}} &44.1 &73.3 &82.5 &2.0 &12.3 &199.9 &43.3 &73.3 &82.5 &2.0 &9.5 &199.1 \\
        {MuLTI\cite{MULTI}} &45.8 & 73.5& 82.0& -& -& 201.3 &- &  -&  -&  -&  -&  - \\ 
        {CLIP2TV\cite{CLIP2TV}} & 46.1  &72.5 &82.9 &2.0  &15.2 &201.5  &43.9 &73 &82.8 &2.0  &11.1 &199.7 \\ 
        {XPool\cite{X_Pool}} &46.9  &72.8 &82.2 &2.0  &14.3 &201.9 &- &- &- &- &- &- \\  
        {DRL$^\dag$\cite{DRL}} &45.7& 73.4& 83.1& 2.0& 13.2& 202.2 &46.3& 72.7& 82.5& 2.0& 9.5& 201.5\\ 
        {STAN\cite{STAN}} &46.9 & 72.8& 82.8& 2.0& -& 202.5 &- &  -&  -&  -&  -&  - \\ 
        {XCLIP$^\dag$\cite{X_CLIP}} &47.4 & 73.4& 83.1& 2.0& 13.7& 203.9 &46.7 &  72.7&  83.0&  2.0&  10.0&  202.4 \\
        {TS2-Net$^\dag$\cite{TS2_Net}} &47.2  &73.7 &83.1 &2.0  &13.1 &204.0 &44.8  &74.3 &84.0 &2.0  &9.3  &203.1  \\ 
        {CAMoE\cite{CAMoE}} &44.6 &72.6 &81.8 &2.0 &13.3 &199.0 &45.1 &72.4 &81.1 &2.0 &10.0 &198.6\\
        {Ours}      &45.3 &74.9 &84.2 &2.0  &12.5 &204.4  &44.8 &74.8 &83.7 &2.0  &9.1  &203.3  \\ \toprule

        {CAMoE$^\ast$\cite{CAMoE}} &47.3 &74.2 &84.5 &2.0 &11.9 &206.0 &49.1 &74.3 &84.3 &2.0 &9.9 &207.7\\
        {Ours$^\ast$} &\textbf{48.5}  &\textbf{77.3}  &\textbf{86.5}  &2.0  &\textbf{11.0}  &\textbf{212.3} &\textbf{49.9}  &\textbf{77.2}  &\textbf{85.8}  &2.0  &\textbf{8.2} &\textbf{212.9}  \\ 

       \bottomrule
     \end{tabular} }
    \vspace{-0.3cm}
\end{table*}
    \begin{table}[]
    \centering
    \caption{Text-to-video retrieval comparison with state-of-the-art methods on DiDeMo.} 
    \vspace{-0.2cm}
    \label{tab:res_didemo}
    \renewcommand{\arraystretch}{0.8}
    \setlength\tabcolsep{2pt}
    \begin{tabular}{l| c c c c c c}
        \toprule
        {Method} & {R$@$1$\uparrow$} & {R$@$5$\uparrow$} & {R$@$10$\uparrow$} & {MdR$\downarrow$} & {MeanR$\downarrow$} & {Rsum$\uparrow$} \\ \toprule
        {CLIP4Clip} & 42.5  &70.2   &80.6   &2.0    &17.5   &193.3 \\ 
        {CLIP2TV} & 45.5  &69.7   &80.6   &2.0    &17.1   &195.8 \\ 
        {DRL$^\dag$} & \textbf{46.5} & 73.9& 83.5& 2.0& \textbf{13.3} & 203.9 \\
        {TS2-Net$^\dag$} & 41.5& 70.9& 80.6& 2.0& 13.9& 193.0 \\ 
        {Ours}          &45.6  &\textbf{75.6} &\textbf{83.6}  &2.0  &13.6  &\textbf{204.8} \\ 
\bottomrule
    \end{tabular}
\end{table}
    \begin{table}[]
    \centering
    \caption{Text-to-video retrieval comparison with state-of-the-art methods on ActivityNet.} 
    \vspace{-0.2cm}
    \label{tab:res_activitynet}
    \renewcommand{\arraystretch}{0.8}
    \setlength\tabcolsep{2pt}
    \begin{tabular}{l| c c c c c c}
        \toprule
         {Method} & {R$@$1$\uparrow$} & {R$@$5$\uparrow$} & {R$@$10$\uparrow$} & {MdR$\downarrow$} & {MeanR$\downarrow$} & {Rsum$\uparrow$} \\ \toprule
        {CLIP4Clip} &40.5   &72.4   &-  &2.0    &7.5    &-  \\  
        {DRL} & \textbf{44.2}   & 74.5  & \textbf{86.1} & 2.0   & - & 204.8 \\     
        {TS2-Net$^\dag$} & 39.9 & 72.3&  84.3&  2.0&  8.5&  196.5 \\ 
        {Ours}  &43.9 &\textbf{75.3} &\textbf{86.1} &2.0 &\textbf{7.1}  &\textbf{205.3} \\
        \bottomrule
    \end{tabular}
  \vspace{-0.2cm}
\end{table}

    Then, to calculate the similarity for each layer, we employ a learnable weighted sum approach. 
    Logically, various actions possess different importance within the overall text, and disparate entities also carry distinct significance for the same action. 
    Therefore, we assign text-video node pairs with different learnable weights $\Vec{w}^{\theta}_{i}$ based on the syntax hierarchy. Specifically, for the whole node $h^{1}_{1}$ in $H^{1}$, we directly set its weight $\Vec{w}^{1}_{i}$ to 1. Besides, for action-level nodes, we utilize the similarity between the overall node $\Vec{h}_{1}^{1}$ and each verb node in $\Vec{H}^{2}$ to assign weights $w^{2}_{i}$ to each action node as follows:
    \begin{equation} 
    \Vec{m}^{t2}_{i}=\operatorname{Norm}\left(\Vec{e}^{\rm{t2}}_{i} + \operatorname{MLP}_{5}\left(\Vec{e}^{\rm{t2}}_{i} \right)\right),
    \end{equation}
    \begin{equation} 
    {sim}^{2}_{i}=\Vec{e}^{\rm{t1}}_{1} \odot \Vec{m}^{\rm{t2}}_{i},
    \end{equation}
    \begin{equation} 
    \Vec{w}^{2}_{i}=\tfrac{\operatorname{exp}\left({sim}^{2}_{i}\right)}
    {\sum_{j\in [1,|\Vec{H}^{2}|]}\operatorname{exp}\left({sim}^{2}_{j}\right)},
    \end{equation}
    where ${sim}^{2}_{i}$ represents the similarity between the overall node and the verb node $h^{2}_{i}$, and $w^{2}_{i}$ is the weight obtained after softmax normalization.

    For the entity noun layer, the degree of association between entity nouns and actions affects the weight. At the same time, the importance of different actions is also introduced into the weight calculation as follows:
    \begin{equation} 
    {sim}^{3}_{i}=\Vec{m}^{\rm{t2}}_{\delta^{3}_{i}} \odot \Vec{e}^{\rm{t3}}_{i},
    \end{equation}
    \begin{equation} 
    \Vec{w}^{3}_{i}=\tfrac{\operatorname{exp}\left({sim}^{2}_{\delta^{3}_{i}}+{sim}^{3}_{i}\right)}
    {\sum_{j\in [1,|\Vec{H}^{3}|]}\operatorname{exp}\left({sim}^{2}_{\delta^{3}_{j}}+{sim}^{3}_{j}\right)},
    \end{equation}
    where $\delta^{\theta}_{i}$ represents the node in the $(\theta-1)$-th layer adjacent to node $\Vec{h}^{\theta}_{i}$ in the $\theta$-th layer.

    So far, based on the three-layer text hierarchy structure and corresponding visual hierarchy structure, we assign weights to each node layer by layer and obtain the comprehensive similarity score of each layer. Finally, we aggregate the similarity score of each layer with the same weight to obtain the final text-guided text-video similarity $\Vec{score}^{\rm{final}}$ as follows:
    \begin{equation} 
    \setlength{\abovedisplayskip}{2pt}
    {score}^{\theta}=\sum_{i\in[1,|\Vec{H}^{\theta}|]}\Vec{w}^{\theta}_{i}\cdot{score}^{\theta}_{i},
    \end{equation}
    \begin{equation} 
    \setlength{\abovedisplayskip}{0pt}
    \setlength{\belowdisplayskip}{2pt}
    {score}^{\rm{final}}=\tfrac{1}{3}\left({score}^{1}+{score}^{2}+{score}^{3}\right),
    \end{equation}
    where $\theta \in \left[1, 2, 3\right]$ represents the level in the syntax hierarchy, and $|\Vec{H}^{\theta}|$ represents the number of nodes in the $\theta$-th layer.

    \begin{table}[]
    \centering
    \caption{Text-to-video retrieval comparison with state-of-the-art methods on MSVD.} 
    \vspace{-0.2cm}
    \label{tab:res_msvd}
    \renewcommand{\arraystretch}{0.8}
    \setlength\tabcolsep{2pt}
    \begin{tabular}{l| c c c c c c}
        \toprule
        {Method} & {R$@$1$\uparrow$} & {R$@$5$\uparrow$} & {R$@$10$\uparrow$} & {MdR$\downarrow$} & {MeanR$\downarrow$} & {Rsum$\uparrow$} \\ \toprule
        {CLIP4Clip} & 45.2  &75.5   &84.3   &2.0    &10.3   &205.0 \\ 
        {CLIP2TV} & 47.0    &76.5   &85.1   &2.0    &10.1   &208.6 \\ 
        {DRL$^\dag$} & 46.5&  76.3&  85.0&  2.0& 10.7& 207.8 \\ 
        {TS2-Net$^\dag$} & 44.0 &75.5   &84.6   &2.0    &10.4   &204.1 \\
        {Ours} &\textbf{47.6}  &\textbf{76.8} &\textbf{85.5}  &2.0  &\textbf{9.4}  &\textbf{209.9} \\ 
        \bottomrule
    \end{tabular}
  \vspace{-0.5cm}
\end{table}
    
\subsection{Loss Calculation and Text-Video Matching}
    Similar to previous methods, for each training step with $B$ text-video pairs, we employ symmetric cross-entropy loss as our training objective function, which is formulated as follows:
    \begin{equation} 
    \setlength{\abovedisplayskip}{2pt}
    \setlength{\belowdisplayskip}{0pt}
    \mathcal{L}^{t2v}=-\tfrac{1}{B}\sum^{B}_{i}log\tfrac{\operatorname{exp}\left(\tau\cdot{score}^{\rm{final}}\left(t_i,v_i\right)\right)}
    {\sum^{B}_{j=1}\operatorname{exp}\left(\tau\cdot{score}^{\rm{final}}\left(t_i,v_j\right)\right)},
    \end{equation}
    \begin{equation} 
    \setlength{\abovedisplayskip}{0pt}
    \setlength{\belowdisplayskip}{0pt}
    \mathcal{L}^{v2t}=-\tfrac{1}{B}\sum^{B}_{i}log\tfrac{\operatorname{exp}\left(\tau\cdot{score}^{\rm{final}}\left(t_i,v_i\right)\right)}
    {\sum^{B}_{j=1}\operatorname{exp}\left(\tau\cdot{score}^{\rm{final}}\left(t_j,v_i\right)\right)},
    \end{equation}
    \begin{equation} 
    \setlength{\abovedisplayskip}{0pt}
    \setlength{\belowdisplayskip}{2pt}
    \mathcal{L}=\tfrac{1}{2}\left(\mathcal{L}^{t2v}+\mathcal{L}^{v2t}\right),
    \end{equation}
    where $\tau$ is the scaling hyper-parameter. During inference, we calculate the similarity score between each text-video pair, and retrieve the most relevant videos with the top-k scores.

\section{EXPERIMENTS}
\subsection{Datasets}
    To demonstrate the effectiveness of our model, we evaluated it on popular benchmark datasets, including MSR-VTT~\cite{MSRVTT}, DiDeMo~\cite{DIDEMO}, ActivityNet\cite{ACTIVITY} and MSVD~\cite{MSVD}. These datasets comprise videos collected across various scenarios, feature varying lengths, and are accompanied by a range of caption quantities.
    \begin{table}[]
    \centering
    \caption{Retrieval performance with different settings of using syntax hierarchy on the MSR-VTT.}
    \vspace{-0.2cm}
    \label{tab:check1}
    \renewcommand{\arraystretch}{0.8}
    \setlength\tabcolsep{1pt}
    \begin{tabular}{l|c|c| c c c c}
        \toprule
        {Method}&{text-syntax}&{video-syntax}& {R$@$1$\uparrow$} & {R$@$5$\uparrow$} & {R$@$10$\uparrow$} & {Rsum$\uparrow$} \\ \toprule
        
        Ours-var1&{-}&{-}&45.4 &73.3 &83.1 &201.9 \\   
        Ours-var2&{\checkmark}&{-}&\textbf{45.8} &73.1 &83.3 &202.2 \\   
        Ours-var3&{\checkmark}&{\checkmark}&45.4 &\textbf{74.6} &\textbf{83.8} &\textbf{203.8}\\
        \bottomrule
    \end{tabular}
    \vspace{-0.3cm}
\end{table}
%
    \begin{table}[]
    \centering
    \caption{Retrieval performance with different settings of text and video syntax hierarchy building on the MSR-VTT.}
    \vspace{-0.2cm}
    \label{tab:check2}
    \renewcommand{\arraystretch}{0.8}
    \setlength\tabcolsep{1pt}
    \begin{tabular}{l|c|c|c|c|c c c c c c}
        \toprule
        {No.}&{Base}&{Verb}&{Noun}&{Adj}& {R$@$1$\uparrow$} & {R$@$5$\uparrow$} & {R$@$10$\uparrow$} & {Rsum$\uparrow$} \\ \toprule
        E1& \checkmark & - & -&-            &44.6 &73.3 &82.7 &200.6 \\ 
        E2& \checkmark & \checkmark & -&-        &45.2 &73.1 &82.7 &201.0 \\   
        E3& \checkmark & \checkmark & \checkmark &-     &44.8 &\textbf{74.7} &\textbf{83.8} &203.3 \\   %
        E4& \checkmark & \checkmark & \checkmark &\checkmark &\textbf{45.4} &74.6 &\textbf{83.8} &\textbf{203.8} \\   
        \bottomrule
    \end{tabular}
    \vspace{-0.3cm}
\end{table}
%

\vspace{-0.15cm}
\subsection{Evaluation Metrics}
    We assessed the performance of various models using standard text-video retrieval metrics: Recall at Rank K (R@K, where higher is preferable), Median Rank (MdR, where lower is preferable), Mean Rank (MeanR, where lower is preferable), and Rsum (where higher is preferable). R@K measures the percentage of correctly retrieved videos within the top K results. In line with previous studies~\cite{CLIP4clip}, we evaluated models across different datasets using R@1, R@5 and R@10. Additionally, echoing the prior research~\cite{TS2_Net}, we aggregated all the R@K results into Rsum to reflect the overall retrieval performance. MdR indicates the median rank of correct videos in the retrieval results, while MeanR computes the average rank of correct videos across the retrieval results.

\vspace{-0.2cm}
\subsection{Implementation Details}
    We initialized the text and visual encoders with pre-trained weight from CLIP (ViT-B/32)~\cite{CLIP}, whereas other modules were initialized randomly. Following the previous studies~\cite{TS2_Net, CLIP4clip}, we set the maximum query text length $N_t$ to 32 and the maximum video frame length $N_v$ to 12 for MSR-VTT and MSVD. For DiDeMo and ActivityNet, we set the maximum query text length $N_t$ and the maximum frame length $N_v$ to 64. The number of layers for the text encoder, visual encoder, and cross encoder are 12, 12, and 4, respectively. The the text embedding and frame embedding dimensions are both set to 512. Additionally, through parameter search, we set the hyperparameter $\lambda_{\rm{frame}}$ to 2 and $\lambda_{\rm{patch}}$ to 4 in our experiments, and the trends in model performance as related to hyperparameter values are presented in subsequent sections. Empirically, following the previous studies \cite{TS2_Net, DRL}, we set the scaling hyperparameter $\tau$ to 4. We utilized the Adam~\cite{ADAM} optimizer to train our model, and due to GPU memory constraints, we used a batch size of 96. The initial learning rate was set at 1e-7 for the text encoder and visual encoder, while the learning rate for other modules was set at 1e-4. In the training phase, we trained the entire model end-to-end. During the validation and testing phases, we first pre-extracted the original text and video features, stored the constructed text syntax structure and the visual features of different granularities that are to be enhanced and fused by the SHE method, and then performed batch inference to obtain the text-video similarity score.

\vspace{-0.1cm}
\subsection{Comparison with State-of-the-Arts}
    \begin{table}[]
    \centering
    \caption{Retrieval performance with different settings of SHE similarity calculation on the MSR-VTT.}
    \vspace{-0.2cm}
    \label{tab:check3}
    \renewcommand{\arraystretch}{0.8}
    \setlength\tabcolsep{1pt}
    \begin{tabular}{l|c|c|c|c|c c c c c c}
        \toprule
        {No.}&{Base}&{Verb}&{Noun\&Adj}&{Fusion}& {R$@$1$\uparrow$} & {R$@$5$\uparrow$} & {R$@$10$\uparrow$} & {Rsum$\uparrow$} \\ \toprule
        E2& \checkmark & \checkmark & -&-        &45.2 &73.1 &82.7 &201.0 \\   
        E5& \checkmark & \checkmark & -& \checkmark   &\textbf{45.9} &73.4 &83.1 &202.4 \\   
        \midrule
        E4& \checkmark & \checkmark & \checkmark &- &45.4 &74.6 &83.8 &203.8 \\   
        E6& \checkmark & \checkmark & \checkmark &\checkmark  &45.3 &\textbf{74.9} &\textbf{84.2} &\textbf{204.4} \\   
        \bottomrule
    \end{tabular}
    \vspace{-0.3cm}
\end{table}
%
    \begin{table}[]
    \centering
    \caption{Retrieval performance with different $\lambda$ settings of video syntax hierarchy building on the MSR-VTT.}
    \vspace{-0.2cm}
    \label{tab:check4}
    \renewcommand{\arraystretch}{0.8}
    \setlength\tabcolsep{1pt}
    \begin{tabular}{c| c c c c c c}
        \toprule
        {Setting} & {R$@$1$\uparrow$} & {R$@$5$\uparrow$} & {R$@$10$\uparrow$} & {Rsum$\uparrow$} \\ \toprule
        {($\lambda_{\rm{frame}}$:2,$\lambda_{\rm{patch}}$:1)}    &44.8 &74.1 &83.0 &202.0 \\
        {($\lambda_{\rm{frame}}$:2,$\lambda_{\rm{patch}}$:2)}    &\textbf{45.7} &74.1 &83.4 &203.2 \\
        {($\lambda_{\rm{frame}}$:2,$\lambda_{\rm{patch}}$:4)}    &45.4 &\textbf{74.6} &\textbf{83.8} &\textbf{203.8} \\
        {($\lambda_{\rm{frame}}$:2,$\lambda_{\rm{patch}}$:6)}    &45.4 &74.4 &83.3 &203.1 \\
        {($\lambda_{\rm{frame}}$:2,$\lambda_{\rm{patch}}$:8)}    &45.2 &74.4 &82.9 &202.5 \\
        \midrule
        {($\lambda_{\rm{frame}}$:1,$\lambda_{\rm{patch}}$:4)}    &45.3 &73.0 &83.0 &201.3 \\
        {($\lambda_{\rm{frame}}$:2,$\lambda_{\rm{patch}}$:4)}    &\textbf{45.4} &74.6 &\textbf{83.8} &\textbf{203.8} \\
        {($\lambda_{\rm{frame}}$:3,$\lambda_{\rm{patch}}$:4)}    &44.3 &\textbf{74.7} &83.4 &202.4 \\
        {($\lambda_{\rm{frame}}$:4,$\lambda_{\rm{patch}}$:4)}    &44.4 &74.1 &83.1 &201.6 \\
        \bottomrule
    \end{tabular}
  \vspace{-0.3cm}
\end{table}
    We compare our method with recent studies on real-world datasets,  including MSR-VTT, MSVD, DiDeMo, and ActivityNet. To exclude performance influences other than architectures, we retrained several prior methods (indicated by the symbol $\dag$) using the official open-source implementations to mitigate possible confounds(such as feature configurations, computing environments .etc) that may interfere with a fair comparison of these architectures.
    As shown in Tab.~\ref{tab:res_msrvtt}, our model surpasses all baseline approaches in Rsum, the primary metric for evaluating overall retrieval performance, and achieves superior results across most metrics in both retrieval directions. Furthermore, by incorporating DSL post-processing, our model yields additional improvements across all metrics.
    To verify the generality and robustness of our model, we conducted experiments across diverse datasets, including DiDeMo, ActivityNet, and MSVD. Notably, for the DiDeMo dataset, as shown in Tab.~\ref{tab:res_didemo}, although our model slightly trails DRL in R@1 and MeanR, it outperforms all other baselines, including DRL, in R@5. Regarding ActivityNet, shown in Tab.~\ref{tab:res_activitynet}, our model achieves notable improvements in R@5 and MeanR metrics when compared to the baseline models. For the MSVD dataset, as shown in Tab.~\ref{tab:res_msvd}, our model demonstrates superior performance on all metrics, notably achieving an improvement of 1.3 in Rsum over the best-performing baselines. In summary, the experimental results across various datasets confirm the effectiveness of our proposed approach.

\vspace{-0.2cm}
\subsection{Training and Inference Time Consumption}
    During the training phase, our model results in an increase in training time by approximately 15\% compared to the baseline model, TS2Net. The computational processes of the model primarily involve extracting textual and visual features, constructing textual syntactic structures, fusing temporal features derived from frame features, and executing syntax-enhanced visual feature fusion and similarity calculations. The first three components can be pre-computed offline and stored for inference, thus significantly reducing computational time during inference. Consequently, our proposed modules incur an additional 5\% in computational cost during inference relative to that of the baseline model, TS2Net. Our method demonstrates a significant improvement in performance with a modest increase in computational time.

\begin{table}[]
    \centering
    \caption{Comparison of Text-to-Video Retrieval Performance Utilizing Visual Encoders with Varying Parameter Quantities on MSR-VTT.}
    \vspace{-0.2cm}
    \label{tab:check5}
    \renewcommand{\arraystretch}{0.8}
    \setlength\tabcolsep{2pt}
    \begin{tabular}{l| c c c c c c}
        \toprule
        {Our Model} & {R$@$1$\uparrow$} & {R$@$5$\uparrow$} & {R$@$10$\uparrow$} & {MdR$\downarrow$} & {MeanR$\downarrow$} & {Rsum$\uparrow$} \\ \toprule
        {ViT-L-32} &48.5  &77.3  &86.5  &2.0  &11.0  &212.3 \\
        {ViT-L-16} &54.5  &79.5  &88.2  &1.0  &9.3  &222.2 \\
        \bottomrule
    \end{tabular}
  \vspace{-0.4cm}
\end{table}
\vspace{-0.2cm}
\begin{figure}[ht]
    \centering
    \includegraphics[width=0.45\textwidth,height=0.4\textwidth]{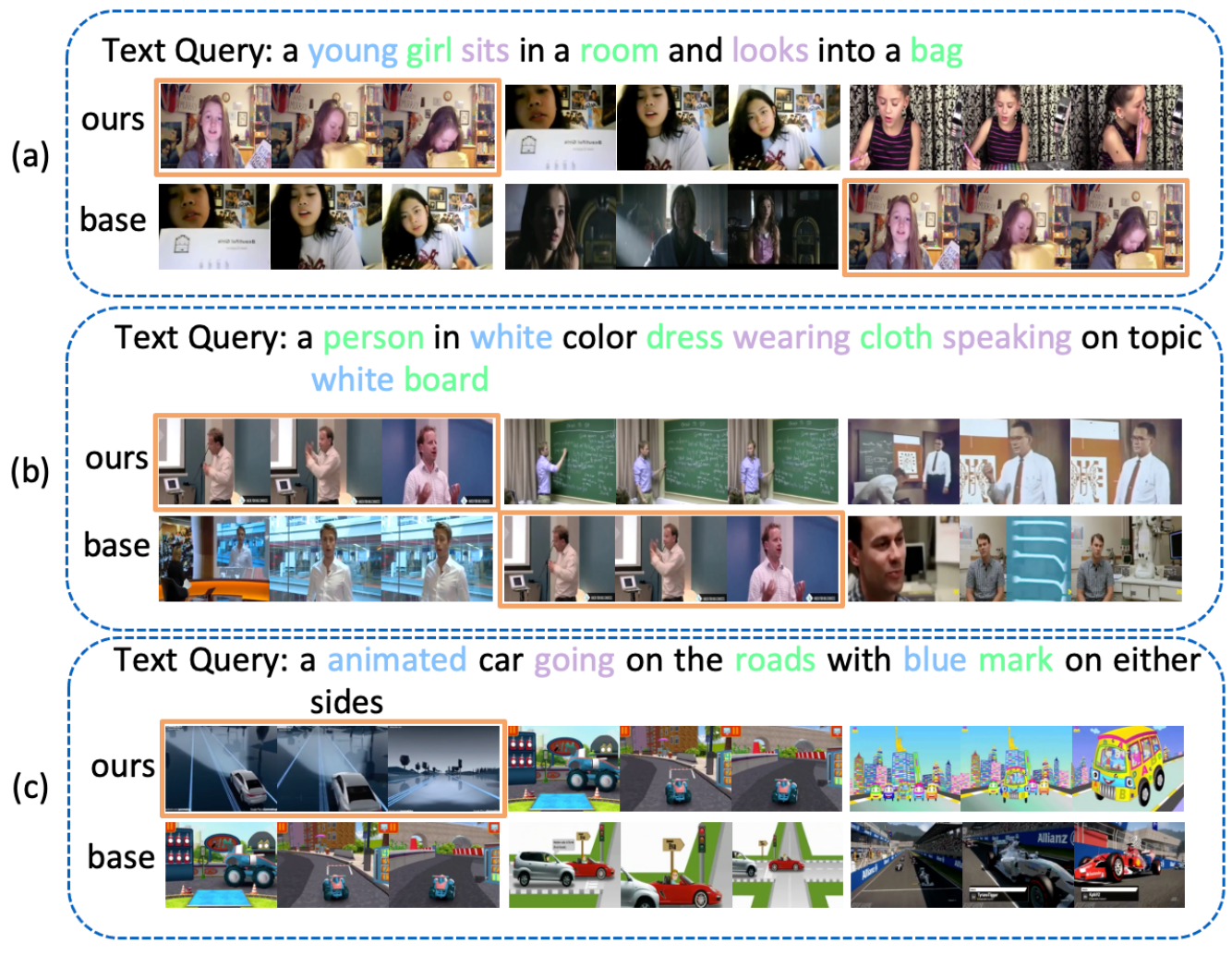} 
    \vspace{-0.4cm}
    \caption{Visualization of text-to-video retrieval results on MSR-VTT. For each query, the top three results are displayed, ranked according to their similarity scores. The upper half of three retrieval groups is generated by our model, whereas the lower half presents results derived by the baseline model. Note that the orange box is the ground truth.}
    \vspace{-0.4cm}
    \label{show_exp}
\end{figure}
\begin{figure}[ht]
    \centering
    \includegraphics[width=0.45\textwidth,height=0.4\textwidth]{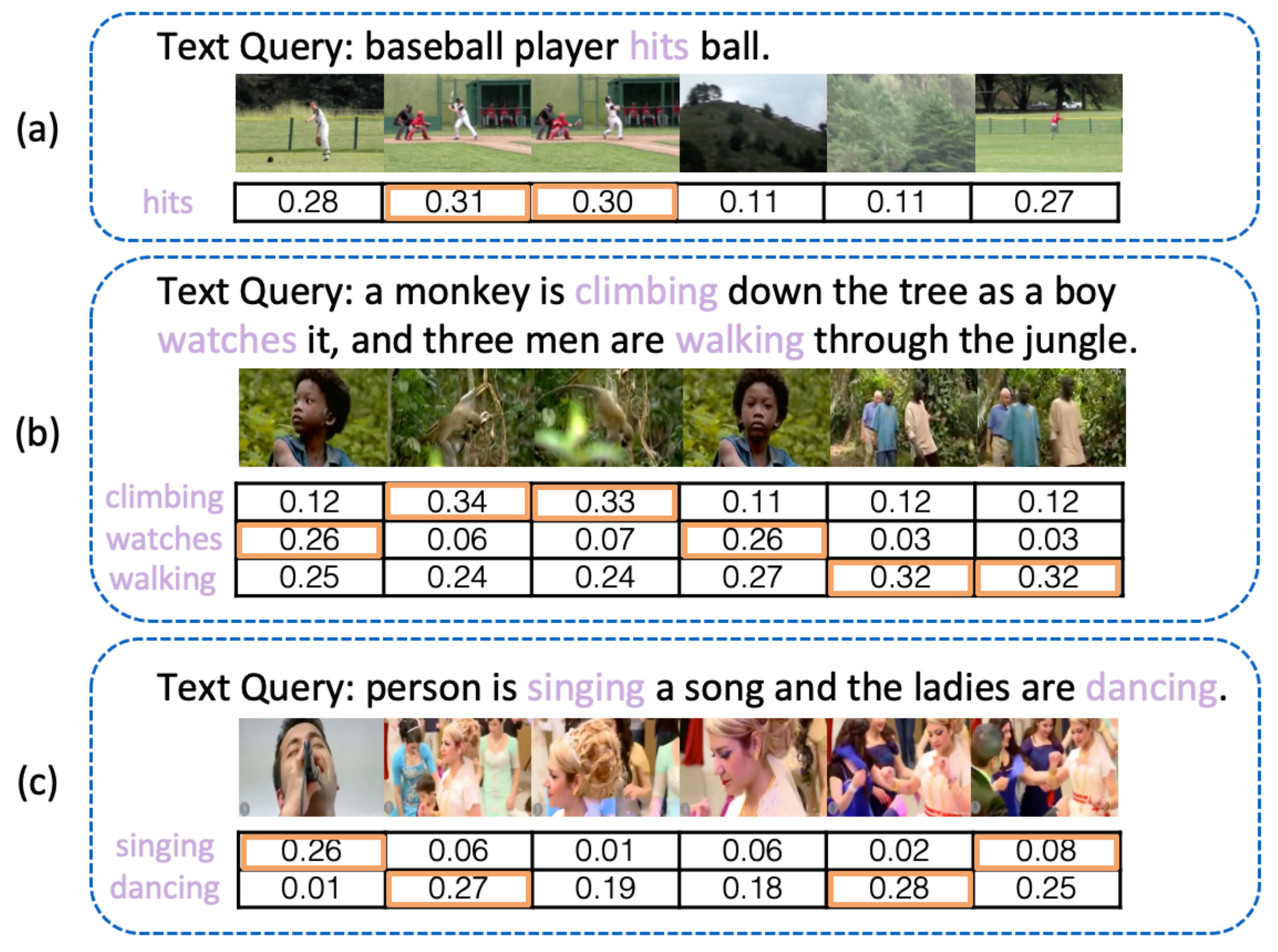} 
    \vspace{-0.4cm}
    \caption{The illustration depicts the construction of the video syntax structure's second layer, known as the verb-action layer. These examples demonstrate the alignment of verbs with frame sequences, highlighting that different verbs or actions are associated with specific frames within the sequence. Our method selects the most $\lambda_{\rm{frame}}$-relevant frames for each verb to construct the visual action features corresponding to the verb.}
    \vspace{-0.4cm}
    \label{show_att}
\end{figure}
\subsection{Ablation Studies}

    \vspace{-0.1cm}
    \subsubsection{Effectiveness of Syntax-Hierarchy-Enhanced Modules}
        To confirm the individual contributions of text and video syntax structures, we modified our framework's text and video SHE feature fusion modules. As illustrated in Tab.\ref{tab:check1}, Ours-var1 indicates that both text and video SHE feature fusion modules are omitted. Ours-var2 indicates that the video syntax feature is excluded and substituted by frame-level CLS features, while Ours-var3 retains the SHE feature fusion module for both modalities. None of the three variations of the adapted models employ the SHE similarity calculation. Tab.~\ref{tab:check1} show that Ours-var2, which utilizes the text syntactic structure, exhibits an improvement in Rsum compared to Ours-var1, which does not utilize any syntactic structure enhancement. Furthermore, Ours-var3 outperforms Ours-var2, confirming the efficacy of our proposed video syntax hierarchy building and feature fusion method.

        To ascertain the effectiveness of our approach in constructing text and video syntax hierarchies, we evaluated the influence of different levels within these hierarchies.
        As indicated in Tab.~\ref{tab:check2}, compared with baseline experiment E1, which utilizes only global text and video features, additionally incorporating each level of the hierarchical structure (\ie{the levels corresponding to verbs (E2), nouns (E3), and adjectives (E4)} ) enhances the model's performance. This suite of experiments confirms that each layer of the four-layer text hierarchy and the three-layer video structure (wherein the textual adjective layer impacts the visual entity layer by augmenting the textual noun layer) is beneficial.

        To validate the SHE similarity calculation method we introduced, we assessed the model's performance by comparing it across different configurations: one using the first two layers and another using all three layers of the enhanced structure. Given that the similarity fusion weight for the first layer is fixed at 1 in our method, a separate comparison for this layer is not required. As indicated in Tab.~\ref{tab:check3}, the SHE similarity calculation method, referred to as Fusion, further improved the performance metrics for both configuration sets.

        Therefore, we concluded that all levels within the syntax hierarchy, along with the SHE similarity calculation method, contribute to the retrieval task, and the various components mutually reinforce each other to yield superior outcomes.

    \vspace{-0.1cm}
    \subsubsection{The Impact of Weights in Video Syntax-Hierarchy Building}
        To investigate the influence of varying weights in constructing the video syntax hierarchy, we conducted a series of experiments by adjusting the weighting parameters $\lambda_{\rm{patch}}$ with $\lambda_{\rm{frame}}$ fixed at 2, and varying $\lambda_{\rm{frame}}$ with $\lambda_{\rm{patch}}$ held constant at 4. Tab.\ref{tab:check4} reveals that the overall retrieval performance initially improves before plateauing (specifically, at $\lambda_{\rm{frame}}$=2, $\lambda_{\rm{patch}}$=4) and subsequently exhibits a slight decline. The primary reason for this trend could be that increasing the parameter values allows for the utilization of more visual information (frames and patches) during the construction phase of the video syntax structure. However, beyond a certain threshold, the negative impact of noise begins to outweigh the positive effects of the effective information, resulting in performance degradation.

    \vspace{-0.1cm}
    \subsubsection{The Impact of Vision Encoders with Different Parameter Sizes}
        As shown in Tab.~\ref{tab:check5}, we verified that employing CLIP with increased parameters can further enhance the performance of our model. For instance, when utilizing CLIP-ViT-L-16 in comparison with CLIP-ViT-L-32, our model exhibits significant improvements across all metrics in the MSRVTT dataset, including R@1$\uparrow$, R@5$\uparrow$, R@10$\uparrow$, MdR$\downarrow$, MeanR$\downarrow$ and Rsum$\uparrow$. In these metrics, our model achieves scores of 54.5 (+6.0), 79.5 (+2.2), 88.2 (+1.7), 1.0 (-1.0), 9.3 (-1.7), and 222.2 (+9.9), respectively.

\subsection{Qualitative Results}
    We visualize several examples from the MSR-VTT dataset for text-to-video retrieval in Fig.\ref{show_exp}. For each query, the top three results are exhibited and ranked according to their similarity scores. The upper half of the three retrieval groups was generated by our full model, whereas the lower half shows results derived from the baseline model. The orange box represents the ground truth. In Fig.\ref{show_exp}(a) and Fig.\ref{show_exp}(b), it can be seen that our model and the baseline are able to retrieve the correct videos; however, our model ranks the correct results higher than the baseline does. Furthermore, in Fig.\ref{show_exp}(c), our model successfully retrieves the video that the baseline model failed to retrieve.

    Additionally, to clarify further the rationale behind the construction of video syntax structure, we present some examples of matching visual features for verbs in Fig.\ref{show_att}. These examples illustrate the alignment of verbs with frame sequences, highlighting that different verbs are associated with distinct frames within the sequence. 
    For instance, in the sentence “A person is singing a song, and the ladies are dancing”, the weights assigned by our methodology to “sing” and “dance” are 0.4970 and 0.5030, respectively, suggesting that these two actions hold remarkably similar importance within this context. Meanwhile, the weights for “person”, “song”, and “ladies” are 0.2970, 0.3543, and 0.3487, respectively, indicating that “song” is more prominent than “person”, while “ladies” also retain significant importance in this context. In this paper, our method retrieves the most $\lambda_{\rm{frame}}$-relevant frames for each verb to construct the visual action features that correspond to the verb, thus mitigating the impact of noise.

\section{Conclusion}
In this paper, to enhance text-video retrieval performance, we propose a novel syntax-hierarchy-enhanced retrieval method that applies the inherent semantic and syntax hierarchy of texts to guide the fusion of visual content and the calculation of text-video similarity. Specifically, we construct a text syntax hierarchy to disclose the grammatical structure of text descriptions, which is subsequently employed to establish the corresponding video syntax hierarchy. We utilize these two syntax hierarchies to improve multi-modal interaction and alignment jointly. Through comprehensive experiments and ablation studies across four datasets, we demonstrate the effectiveness of our proposed method.




\bibliographystyle{IEEEtran}
\bibliography{reference}

\vfill

\end{document}